\documentclass[letterpaper,10pt,conference]{ieeeconf}

\IEEEoverridecommandlockouts
\usepackage{graphicx}
\usepackage{mathptmx}
\usepackage{amsmath,amssymb}
\usepackage{mathtools}
\usepackage{booktabs}
\usepackage{balance}
\usepackage{stfloats}
\usepackage[hidelinks]{hyperref}

\newcommand{\plannedgraphic}[4]{%
  \IfFileExists{figs/#1.pdf}{\includegraphics[width=#2]{figs/#1.pdf}}{%
    \IfFileExists{figs/#1.png}{\includegraphics[width=#2]{figs/#1.png}}{%
      \fbox{\parbox[c][#3][c]{\dimexpr#2-2\fboxsep-2\fboxrule\relax}{%
        \centering\footnotesize\textbf{Figure pending: \texttt{\detokenize{#1}}}\\[3pt]#4}}}}%
}

\title{\LARGE \bf
Volumetric Harmonic Field Navigation for Quadrotors
}

\author{Shuxiu Jia, Amartya Mukherjee, and Yating Yuan
\thanks{This research was supported by Mitacs under the Mitacs Globalink Research Internship award, Application ID 209332, and by the Waterloo.AI Physical AI Graduate Scholarship, supported by the Philantra Foundation.}
\thanks{Shuxiu Jia is with the Hong Kong Polytechnic University, Hong Kong (email: {\tt\small shuxiu.jia@connect.polyu.hk}).}
\thanks{Amartya Mukherjee and Yating Yuan are with the Department of Applied Mathematics, University of Waterloo, Waterloo, Ontario, Canada N2L 3G1 (email: {\tt\small (a29mukhe,yating.yuan,j.liu)@uwaterloo.ca}).}
}

\begin{document}
\maketitle

\begin{abstract}
Quadrotor navigation in cluttered 3-D environments requires global guidance while local motion remains subject to collision and motion limits. Harmonic potentials provide dense guidance from a global boundary value problem, but coupling a volumetric harmonic field to constrained physical quadrotor motion remains an open experimental problem. We couple a precomputed volumetric harmonic field with a constrained predictive planner that queries the field at predicted positions instead of extracting a global reference path. In Structured 3-D tests, harmonic guidance yields larger minimum clearance and lower RMS jerk than matched Dijkstra guidance, at the cost of longer paths; the same pattern remains when both methods use the same passage. Long maze tests span routes far beyond one prediction horizon, and Crazyflie trials validate physical execution. To the best of our knowledge, this is the first physical quadrotor demonstration of volumetric harmonic field navigation. The results show that globally constructed harmonic guidance can directly support local constrained motion generation on a physical quadrotor.
\end{abstract}

\section{Introduction}
\label{sec:introduction}

Quadrotor navigation in cluttered 3-D environments requires guidance over the environment and motion that respects constraints. Modern planners combine geometric planning, trajectory optimization, and receding horizon control~\cite{liu2017sfc,zhou2019fastplanner,wang2022gcopter,liu2024ipc}. Harmonic fields offer a different form of guidance: a scalar function defined throughout the free volume.

Potential fields map environment geometry to local motion guidance~\cite{khatib1986apf}. Classical artificial potential fields can contain spurious local minima and trap the robot~\cite{koren1991limitations}. More structured constructions include navigation functions~\cite{rimon1992navigation} and harmonic potentials~\cite{connolly1990laplace,kim1992harmonic}. Harmonic navigation solves Laplace's equation with boundary values on the goal and unsafe boundaries. For a nonconstant harmonic solution, the maximum principle excludes interior local maxima and minima, although saddle points may remain~\cite{connolly1990laplace,connolly1993applications}. The resulting potential is smooth in the interior and defines a dense
scalar field over the solved free domain. Using such a field in 3-D still requires a practical way to turn field queries into constrained motion.

Physical aerial systems have used harmonic or Laplacian guidance~\cite{scherer2008flying,motonaka2015hardware,bassolillo2025swarm}, while volumetric harmonic fields for UAV navigation and exploration have also been studied~\cite{rasche2013uav,kopo2023harmonic3d}. The remaining gap is a documented physical quadrotor realization of volumetric harmonic field navigation. We address it by coupling a volumetric harmonic field to constrained predictive motion generation over a short horizon. To the best of our knowledge, this is the first physical quadrotor demonstration of volumetric harmonic field navigation.

\begin{figure*}[!ht]
  \centering
  \includegraphics[width=\textwidth]{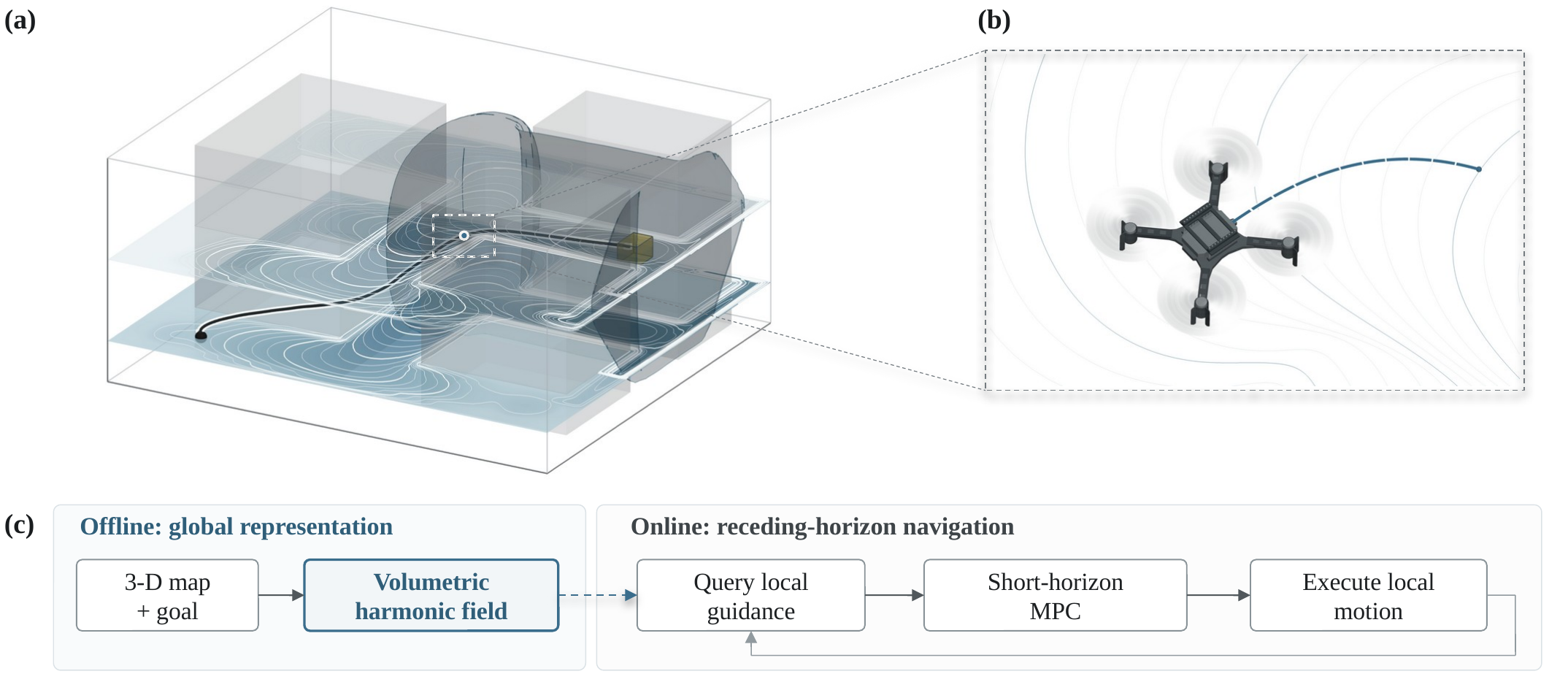}
  \caption{%
  Volumetric harmonic guidance and online predictive navigation.
  \textbf{(a)} Volumetric harmonic field in the known 3-D workspace,
  visualized by two horizontal equipotential sections and one translucent
  3-D isosurface. The black curve is the measured Crazyflie flight from
  start to goal, not a preplanned reference path.
  \textbf{(b)} Local view of the boxed region in (a), showing the Crazyflie
  and the current 12-stage MPC prediction generated online from
  harmonic field guidance.
  \textbf{(c)} System pipeline: the harmonic field is constructed offline
  and queried at each MPC update.}
  \label{fig:teaser}
\end{figure*}

Experiments in structured 3-D scenes, a long maze, and physical Crazyflie flight evaluate the approach. With the same predictive planner, harmonic guidance gives larger minimum clearance and lower RMS jerk than Dijkstra guidance, but follows longer paths. The pattern remains when both methods use the same passage, so coarse route choice alone does not explain it. The long maze tests navigation over routes far beyond one prediction horizon. The first contribution is the physical realization of volumetric harmonic navigation through constrained predictive motion generation. The second is an empirical characterization of the resulting clearance, jerk, and path length behavior.

\section{Related Work}
\label{sec:related_work}

\subsection{Quadrotor Motion Planning}

Quadrotor planners often combine a geometric description of free space with trajectory generation and control. Polynomial trajectories that minimize snap provide smooth references for agile flight~\cite{mellinger2011minsnap}. Safe Flight Corridors represent free space as connected convex regions for constrained trajectory optimization~\cite{liu2017sfc}. Related planning and tracking designs have also used corridor constructions to establish formal reach-avoid guarantees for quadrotors~\cite{serry2026reachavoid}. Fast-Planner combines kinodynamic search with B-spline optimization and time adjustment~\cite{zhou2019fastplanner}; GCOPTER handles multicopter trajectory optimization under geometric and dynamic constraints~\cite{wang2022gcopter}. IPC combines A* planning, Safe Flight Corridors, and MPC with jerk input~\cite{liu2024ipc}. The cited planners organize free space guidance as a path, corridor, or trajectory. Our planner instead queries a volumetric harmonic field at predicted states.

\subsection{Harmonic Navigation}

Khatib introduced artificial potential fields for obstacle avoidance in real time~\cite{khatib1986apf}. Koren and Borenstein later documented limitations including local trapping~\cite{koren1991limitations}. Navigation functions offer one structured construction~\cite{rimon1992navigation}; harmonic potentials offer another through Laplace's equation~\cite{connolly1990laplace,kim1992harmonic}.

Harmonic guidance has also been used on aerial platforms. Masoud used an HPF gradient as a velocity reference for joint UAV planning and control~\cite{masoud2011uav}. Scherer et al. used a 3-D Laplacian global planner with local collision avoidance on an autonomous helicopter~\cite{scherer2008flying}. Rasche developed harmonic-potential-based UAV navigation and exploration for 3-D environments~\cite{rasche2013uav}. Cruz and Encarna\c{c}\~ao used 3-D harmonic panel fields for fixed-wing UAV obstacle avoidance in a hardware-in-the-loop simulation~\cite{cruz2012uav}. Motonaka et al. generated 3-D quadrotor motion from harmonic fields defined on the three coordinate planes and later tested the method on a physical quadrotor~\cite{motonaka2014hpf3d,motonaka2015hardware}. Those experiments used planar fields rather than one volumetric harmonic solution.

Later work broadened harmonic planning and control. Wray et al. introduced harmonic computation in log space and GPU acceleration~\cite{wray2016logspace}. Kopo et al. used an accelerated boundary element method for harmonic exploration in complex 3-D environments and validated it in simulation~\cite{kopo2023harmonic3d}. Recent methods have also extended harmonic potentials to online task adaptation and path class customization~\cite{wang2025hybrid,wang2025customize}. Mukherjee et al. developed harmonic control Lyapunov barrier functions for formal reach-avoid control and included a numerical 2-D quadrotor example~\cite{mukherjee2023hclbf}. Recent physical studies still do not document volumetric harmonic flight. Bassolillo et al. reported CoDrone experiments with Laplace-derived fields, but the published field examples and flight trajectories are planar~\cite{bassolillo2025swarm}. Kotsinis et al. demonstrated harmonic and Poisson field exploration on a physical Crazyflie in complex 2-D spaces~\cite{kotsinis2026harmonic}. The published quadrotor and multirotor hardware studies above are planar or do not document a volumetric harmonic field.

\section{Volumetric Harmonic Navigation}
\label{sec:method}

\subsection{Navigation Task and Planning Model}
\label{subsec:task-dynamics}

Let $\mathcal{W}\subset\mathbb{R}^3$ be a known bounded workspace. Let $\mathcal{O}\subset\mathcal{W}$ be the obstacle set after inflation by the vehicle radius and safety margin. The planning free space is $\mathcal{F}=\operatorname{int}(\mathcal{W})\setminus\mathcal{O}$. Let $\mathcal{G}\subset\mathcal{F}$ be a compact goal region with center $\mathbf{p}_g$, and let $\Omega\subseteq\mathcal{F}$ be the connected component containing $\mathcal{G}$. Starting from $\mathbf{p}_0\in\Omega$, the task is to reach $\mathcal{G}$ without leaving $\Omega$ while satisfying the stated translational motion limits. A volumetric harmonic field supplies guidance over $\Omega$. A planner with a short horizon turns the queried directions into constrained motion.

The predictive planner uses a third-order translational model~\cite{liu2024ipc}
\begin{equation}
\label{eq:planning-dynamics}
\mathbf{x}=\begin{bmatrix}\mathbf{p}^\top&\mathbf{v}^\top&\mathbf{a}^\top\end{bmatrix}^{\!\top},\qquad
\dot{\mathbf{p}}=\mathbf{v},\quad
\dot{\mathbf{v}}=\mathbf{a},\quad
\dot{\mathbf{a}}=\mathbf{j},
\end{equation}
where $\mathbf{p},\mathbf{v},\mathbf{a},\mathbf{j}\in\mathbb{R}^3$ denote position, velocity, acceleration, and jerk. Jerk is the planning input. 
Assuming the jerk input is constant over each sampling interval of
duration $\Delta t$, the exact discrete-time model is
\begin{equation}
\label{eq:discrete-dynamics}
\begin{aligned}
\mathbf{x}_{k+1}&=A_d\mathbf{x}_k+B_d\mathbf{j}_k,\\
A_d&=
\begin{bmatrix}
I_3 & \Delta t I_3 & \frac{1}{2}\Delta t^2 I_3\\
0   & I_3          & \Delta t I_3\\
0   & 0            & I_3
\end{bmatrix},\\
B_d&=
\begin{bmatrix}
\frac{1}{6}\Delta t^3 I_3\\
\frac{1}{2}\Delta t^2 I_3\\
\Delta t I_3
\end{bmatrix},
\end{aligned}
\end{equation}
where $I_3$ is the $3\times3$ identity matrix.
The planner bounds velocity, acceleration, and jerk and imposes collision constraints on position. Attitude dynamics are not part of the planning model. During physical flight, the low-level controller tracks the planned reference and handles attitude stabilization and thrust allocation. Table~\ref{tab:experimental-settings} lists the motion limits used in the experiments.

\subsection{Volumetric Harmonic Field}
\label{subsec:harmonic-guidance}

Let $V:\Omega\setminus\mathcal{G}\rightarrow[0,1]$ be the harmonic potential defined by the Dirichlet problem
\begin{equation}
\label{eq:harmonic-bvp}
\nabla^2 V(\mathbf{p})=0\quad \mathbf{p}\in\Omega\setminus\mathcal{G},\qquad
V|_{\Gamma_g}=0,\qquad V|_{\Gamma_u}=1.
\end{equation}
Here $\Gamma_g=\partial\mathcal{G}$ and $\Gamma_u=\partial\Omega$. The unsafe boundary $\Gamma_u$ contains the inflated obstacle boundaries and the workspace boundary of the connected free space component that contains the goal. We retain the solution as a volumetric navigation field during execution rather than convert it to a global reference path. For a nonconstant harmonic solution, the maximum principle excludes interior local maxima and minima, although saddle critical points may remain~\cite{connolly1990laplace,connolly1993applications}.

At points where $\nabla V(\mathbf{p})\neq 0$, the guidance direction is
\begin{equation}
\label{eq:harmonic-direction}
\mathbf{d}_H(\mathbf{p})=-\frac{\nabla V(\mathbf{p})}{\|\nabla V(\mathbf{p})\|_2}.
\end{equation}
The field is computed offline. We solve the complementary field
$H=1-V$ on a Cartesian voxel grid with spacing $h$, using a seven-point finite-difference stencil and a 3-D log-space Gauss--Seidel iteration adapted from~\cite{wray2016logspace}. Since $\nabla H=-\nabla V$, ascending $H$
is equivalent to descending $V$. For a query between grid nodes, we
locally rescale the eight neighboring log values before evaluating the
trilinear gradient. The scale cancels after normalization. A numerically
vanishing gradient gives zero guidance. During each planning update,
the field is queried only at the nominal predicted positions within the
finite horizon.

\subsection{Predictive Motion Generation}
\label{subsec:field-mpc}

For an $N$-stage horizon, each replanning step shifts the previous jerk
sequence and rolls it out through the discrete model to obtain nominal
positions $\{\bar{\mathbf{p}}_k\}_{k=1}^{N}$.
We query the harmonic field along the nominal rollout,
\begin{equation}
\label{eq:nominal-field-query}
\mathbf{d}_k=\mathbf{d}_H(\bar{\mathbf{p}}_k),\qquad
P_{\perp,k}=I_3-\mathbf{d}_k\mathbf{d}_k^\top,
\end{equation}
The queried directions
remain fixed during one quadratic program (QP) solve. The QP optimizes
the jerk sequence $\{\mathbf{j}_k\}_{k=0}^{N-1}$ with objective
\begin{align}
\label{eq:mpc-objective}
J={}&\sum_{k=1}^{N}\Big[
q_{\parallel}(\mathbf{d}_k^\top\mathbf{v}_k-s_k)^2
+q_{\perp}\mathbf{v}_k^\top P_{\perp,k}\mathbf{v}_k
+q_a\|\mathbf{a}_k\|_2^2\Big] \nonumber\\
&+\sum_{k=0}^{N-1}\Big[
q_j\|\mathbf{j}_k\|_2^2
+q_{\Delta j}\|\mathbf{j}_k-\mathbf{j}_{k-1}\|_2^2\Big]
+\ell_N,
\end{align}
where $\mathbf{j}_{-1}$ is the previously applied jerk and
$q_{\parallel}$, $q_{\perp}$, $q_a$, $q_j$, and $q_{\Delta j}$ are
nonnegative weights. The first term tracks speed $s_k$ along the queried
direction, the second suppresses transverse velocity, and the remaining
stage terms regularize acceleration, jerk, and changes in jerk. Outside
the goal region, $s_k$ is the cruise speed limited by the stopping speed
$\sqrt{2a_{\max}d_{\mathcal G,k}}$, with a $0.18$~m/s floor; it is zero
inside the goal region. Here $d_{\mathcal G,k}$ is the distance from
$\bar{\mathbf{p}}_k$ to $\mathcal G$. The terminal cost $\ell_N$ uses
weights $q_v^N$, $q_a^N$, and $q_p^N$ to penalize velocity error relative
to $s_N\mathbf{d}_N$, terminal acceleration, and position error to
$\mathbf{p}_g$; the position term is activated linearly within $0.75$~m
of the goal region.

The QP enforces the discrete dynamics and polyhedral approximations of
the velocity, acceleration, and jerk bounds. With constant jerk over
each stage, position and velocity are cubic and quadratic in time,
respectively. Their B\'ezier control points are used to impose
workspace, collision, and velocity constraints over the full stage.
Nearby obstacles are represented by supporting half-spaces constructed
about the nominal rollout, and a trust region limits deviations from
that rollout. The resulting online problem is a convex QP solved with
OSQP~\cite{stellato2020osqp}. A separate continuous-time validator
checks the candidate trajectory for workspace, collision, velocity,
acceleration, and jerk constraints. If accepted, the first two stages
are committed before replanning from the latest state.

\section{Experiments}
\label{sec:experiments}

Structured 3-D tests characterize navigation behavior in a volumetric scene. A matched Dijkstra comparison isolates differences in guidance; the same passage subset checks whether those differences persist without a coarse passage change. The large maze tests routes far beyond the 0.60~s prediction horizon. Crazyflie trials evaluate the same harmonic guidance and predictive planning chain in flight.

\begin{figure*}[!t]
  \centering
  \includegraphics[width=\textwidth]{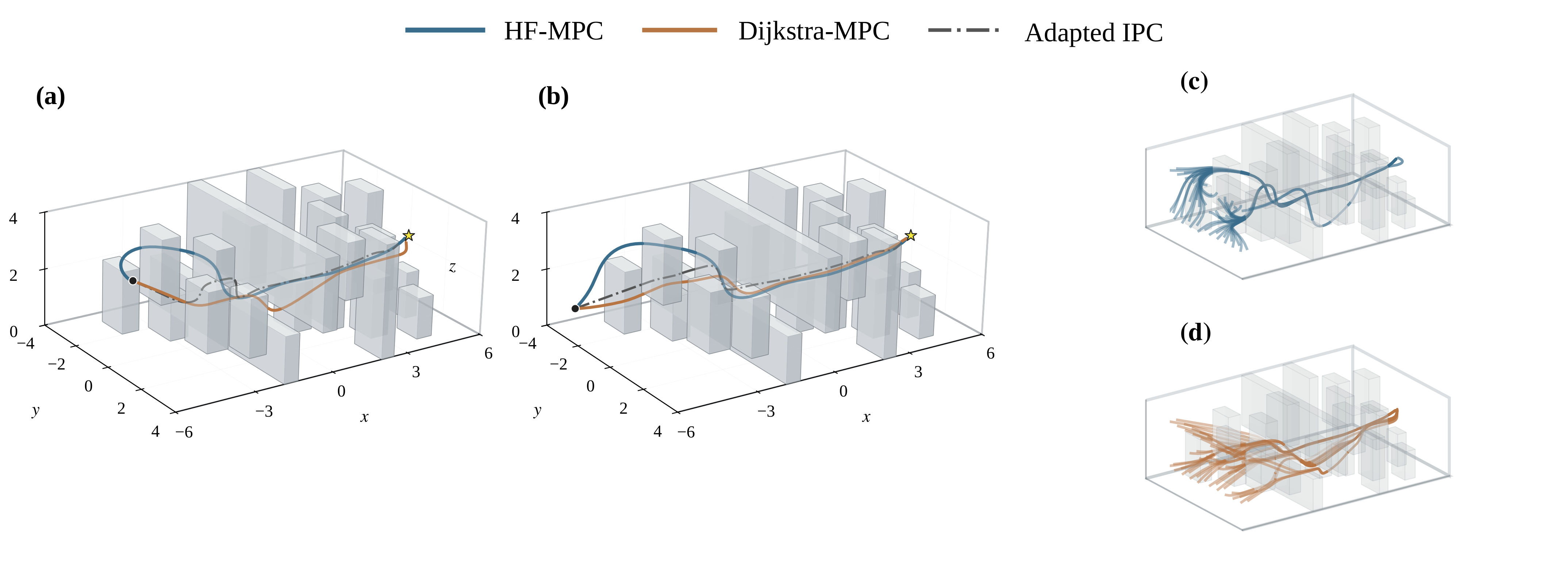}
  \caption{Structured 3-D trajectories. (a) HF-MPC and Dijkstra-MPC take different passage classes. (b) They take the same passage class while remaining geometrically distinct. Adapted IPC is shown for system context. (c) All 50 HF-MPC trajectories. (d) All 50 Dijkstra-MPC trajectories.}
  \label{fig:structured-results}
\end{figure*}

\subsection{Experimental Setup}
\label{subsec:experimental-setup}

For the controlled comparison, Dijkstra-MPC uses the same grid, planner, constraints, solver, execution policy, and success criterion as HF-MPC. A reverse 26-connected Dijkstra search computes the cost from every reachable node to the goal using Euclidean edge costs and no diagonal corner cutting. Each node stores a minimum cost descent direction. Interpolation gives directions between grid nodes, and the planner queries them at the same predicted positions used for harmonic guidance. Grid-based goal cost guidance has prior quadrotor use~\cite{wang2021decentralized}, and interpolated grid navigation functions have been coupled directly to constrained MPC for wheeled robots~\cite{seder2024convergent}; the comparator here is constructed specifically on the shared HF grid and planner.

Adapted IPC serves as a system reference~\cite{liu2024ipc}. A* first produces a path without collisions. We simplify that path and cover it with local convex Safe Flight Corridors. A linear MPC with jerk input tracks the path inside the active corridor. For our known static scenes, we replace the LiDAR map with the same analytic map and build the corridors from known obstacle geometry. We use common vehicle limits and task criteria while retaining IPC's $0.10$~s prediction step, $N=15$ horizon, and 100~Hz backend. Attitude and thrust conversion are outside the translational benchmark. The comparison with IPC is a system comparison, not a controlled change of guidance alone.

All methods share the same map, start, goal, vehicle limits, timeout,
and evaluation rules. Success requires entering the goal region without
collision and maintaining speed at most 0.3~m/s and acceleration at most
0.5~m/s$^2$ for 0.3~s. For each successful simulated episode of duration
$T$, path length is the arc length of the continuous trajectory, minimum
clearance is the minimum vehicle-body clearance from any obstacle or
workspace boundary, and RMS jerk is
$j_{\rm RMS}=\left(T^{-1}\int_0^T\|\mathbf{j}(t)\|_2^2\,dt\right)^{1/2}$.
Table~\ref{tab:experimental-settings} gives the HF-MPC and Dijkstra-MPC
planning settings.

\begin{table}[!ht]
\caption{HF-MPC and Dijkstra-MPC planning settings.}
\label{tab:experimental-settings}
\centering
\footnotesize
\setlength{\tabcolsep}{3.4pt}
\begin{tabular}{@{}lc@{}}
\toprule
Setting & Value \\
\midrule
$N$ / $\Delta t$ / prediction horizon & $12$ / $0.05$~s / $0.60$~s \\
Replanning rate / QP solver & $10$~Hz / OSQP \\
$(v_{\max},a_{\max},j_{\max})$ &  $(1.4,2.5,6.0)$ \\
$(q_{\parallel},q_{\perp},q_a,q_j,q_{\Delta j})$ & $(8,4,0.15,0.04,0.12)$ \\
Terminal $(q_v^N,q_a^N,q_p^N)$ & $(2,0.5,3)$ \\
$h$ (Structured / maze / flight) & $0.08$ / $0.30$ / $0.04$~m \\

\bottomrule
\end{tabular}
\end{table}

Both guidance fields are built offline. On an Apple M1, median build times using one thread were 535 and 127~s for HF and Dijkstra in Structured 3-D, and 67.4 and 28.1~s in the large maze; file I/O is excluded.

\subsection{Structured 3-D Evaluation}
\label{subsec:structured-results}

\textit{Passage choice.} All three methods completed all 50 starts. Passage class records only the designated lateral opening crossed at the Zone D gate. HF-MPC and Dijkstra-MPC took different passages in 23 starts. Dijkstra-MPC split evenly between the two openings; HF-MPC used the opening at negative $y$ in 48 starts. Figure~\ref{fig:structured-results}(a) shows one different passage case. Table~\ref{tab:structured-results} reports median metrics for all three methods. The IPC values are included for system context, not as a matched guidance comparison.

\begin{table}[!ht]
\caption{Structured 3-D median metrics over 50 successful starts.}
\label{tab:structured-results}
\centering
\footnotesize
\setlength{\tabcolsep}{2.6pt}
\begin{tabular}{@{}lrrr@{}}
\toprule
& \shortstack{Path\\length [m]} & \shortstack{Min.\\clearance [m]} & \shortstack{RMS jerk\\{[m/s$^3$]}} \\
\midrule
HF-MPC & 13.67 & 0.254 & 1.355 \\
Dijkstra-MPC & 11.71 & 0.056 & 1.775 \\
Adapted IPC & 12.49 & 0.248 & 2.078 \\
\bottomrule
\end{tabular}
\end{table}

\begin{figure}[!ht]
  \centering
  \includegraphics[width=\columnwidth,height=1in]{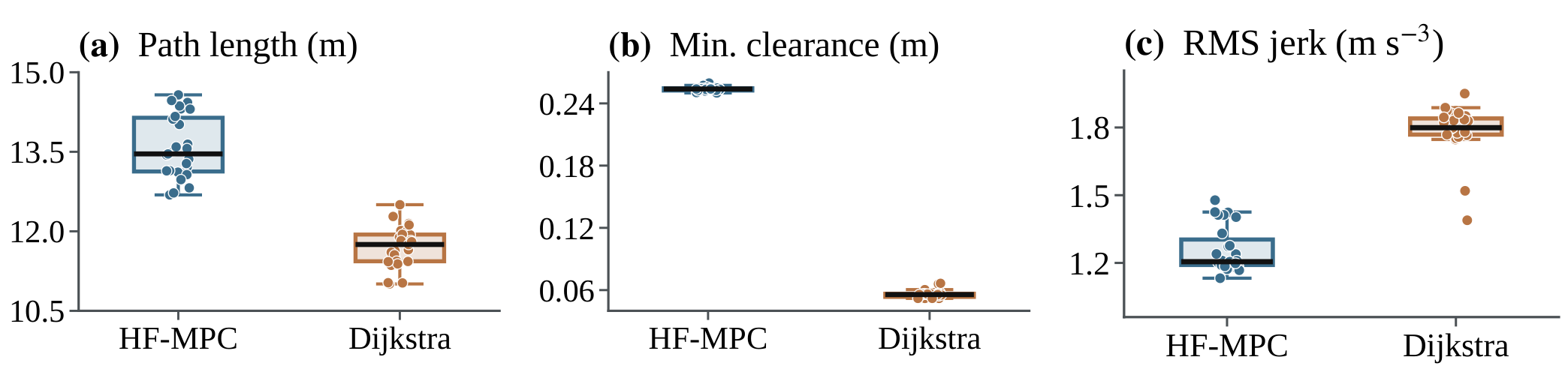}
  \caption{Metrics for the 27 same passage starts. Each point is one deterministic start. Boxes show the median and interquartile range with 1.5 IQR whiskers. Paired differences are reported in the text.}
  \label{fig:same-passage-metrics}
\end{figure}

\textit{Same passage behavior.} We next retain the 27 starts where HF-MPC and Dijkstra-MPC crossed the same designated opening. Figure~\ref{fig:same-passage-metrics} shows the absolute metric distributions. Across all 27 pairs, HF-MPC follows a longer path, has larger minimum clearance, and has lower RMS jerk. For metric $m$ and matched start $i$, let $\Delta m_i=m_i^{\mathrm{HF}}-m_i^{\mathrm{Dij}}$. The median $[Q_1,Q_3]$ differences are $+1.95\,[1.42,2.26]$~m for path length, $+0.198\,[0.196,0.201]$~m for minimum clearance, and $-0.484\,[-0.533,-0.336]$~m/s$^3$ for RMS jerk. The same ordering remains after the passage class is fixed. Figure~\ref{fig:structured-results}(b) shows the same passage case; panels (c) and (d) show all 50 trajectories.

\subsection{Route Scale and Prediction Horizon}
\label{subsec:maze-results}

\begin{figure}[!ht]
  \centering
  \includegraphics[width=\linewidth]{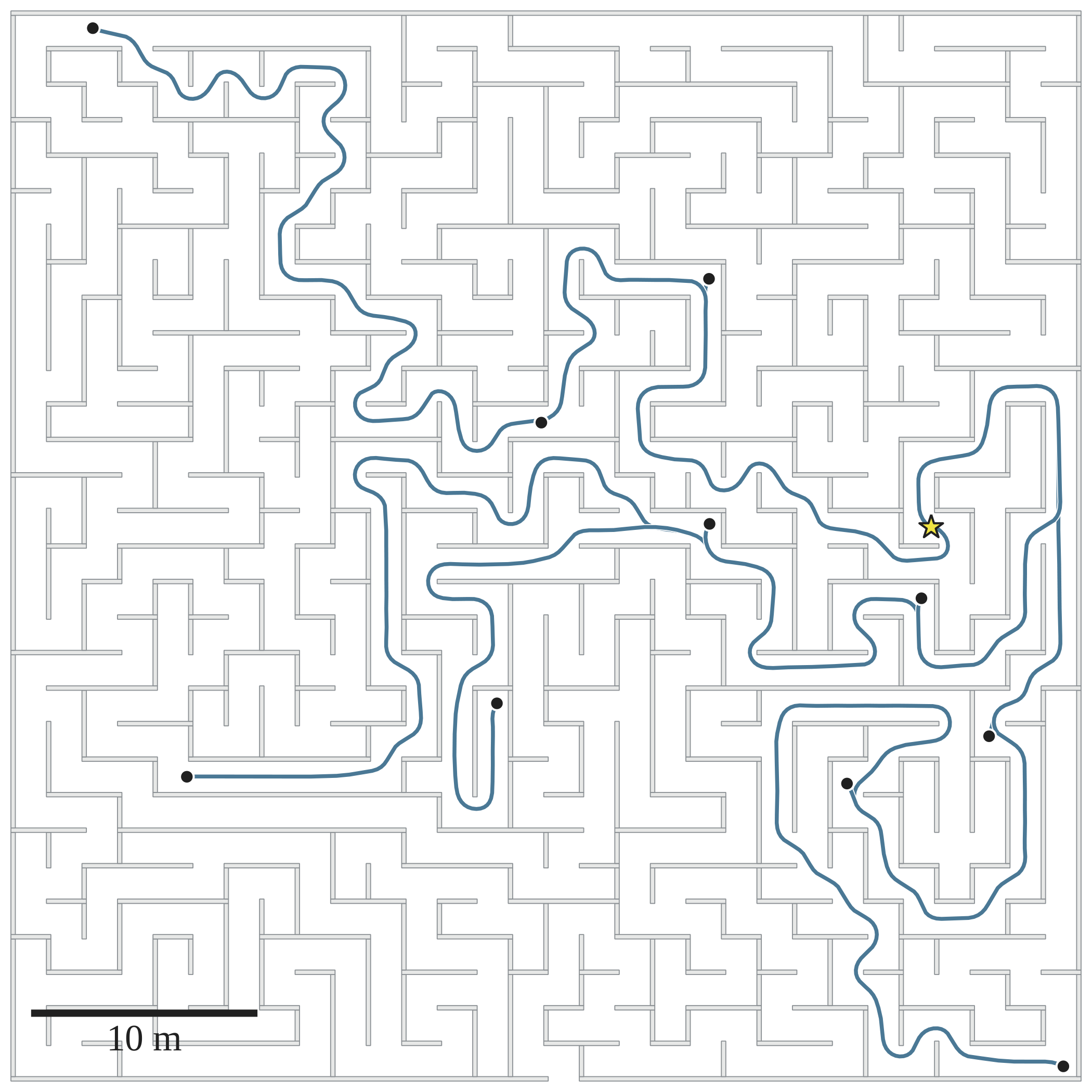}
  \caption{Large maze HF-MPC trajectories from 10 fixed starts to a common goal.}
  \label{fig:maze-validation}
\end{figure}

The large maze tests navigation on routes much longer than the local prediction horizon. We use the maze map from~\cite{wray2016logspace} and fix 10 starts with a common goal. Reference route lengths range from 28.4 to 92.8~m. At 0.95~m/s, one 0.60~s horizon covers about 0.57~m; the routes are about 50 to 163 times longer. The horizon remains fixed, and HF-MPC completed all 10 starts. Figure~\ref{fig:maze-validation} shows the trajectories over the full maze.

\subsection{Physical Flight}
\label{subsec:hardware-results}

The flight trials use the deployed $h=0.04$~m harmonic field. OptiTrack/Motive supplies pose measurements through Crazyswarm~\cite{preiss2017crazyswarm}. Offboard HF-MPC replans at 10~Hz, and a separate process streams validated references at 100~Hz to the onboard Crazyflie controller.

\begin{figure*}[!ht]
  \centering
  \includegraphics[width=\linewidth]{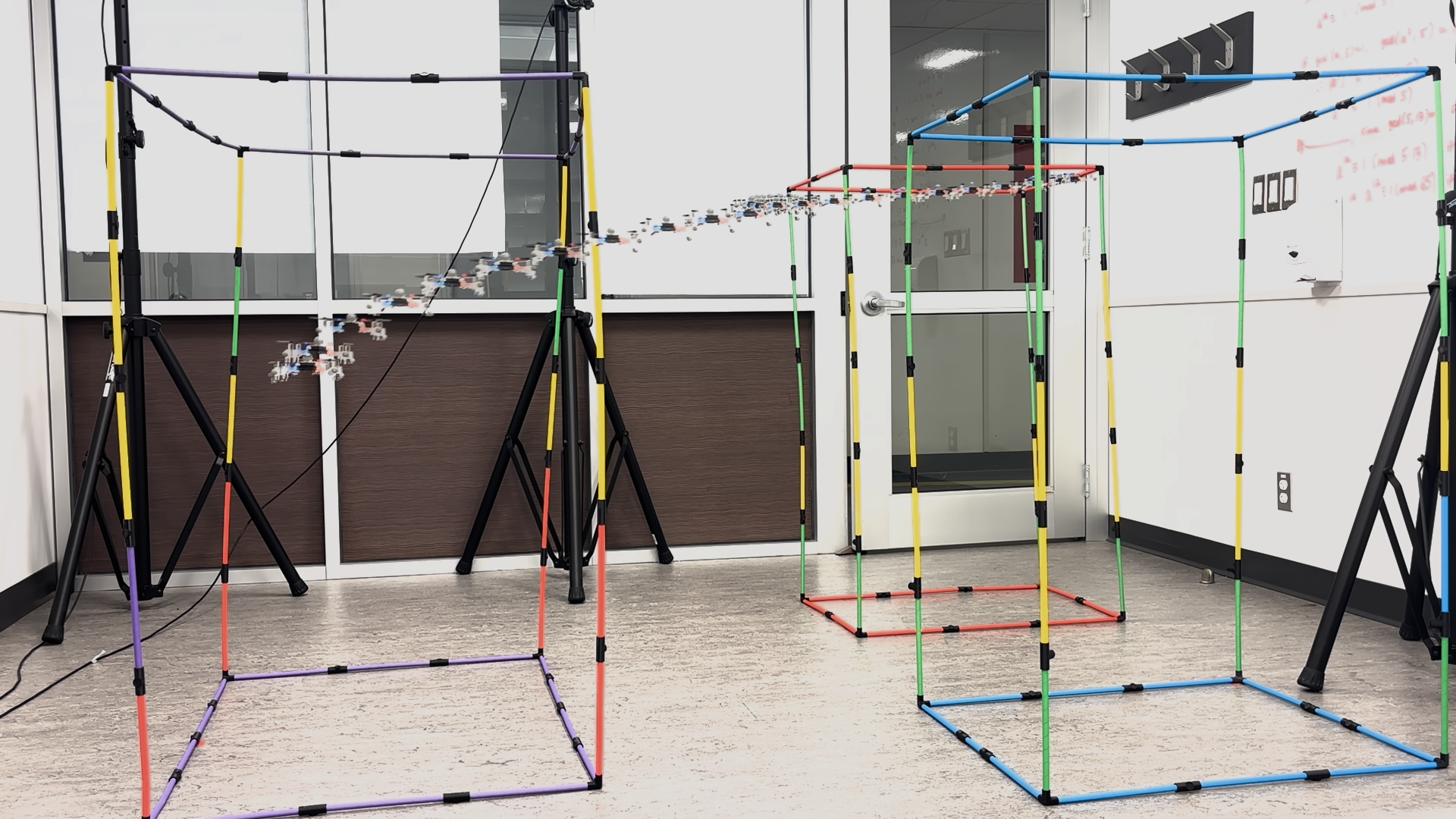}
  \caption{Crazyflie execution of HF-MPC in the motion capture arena. The quadrotor navigates from start to goal while avoiding the obstacles. The straw-frame boxes represent obstacles in the environment. The overlaid sequence shows the measured flight.}
  % \caption{Crazyflie execution of HF-MPC in the motion capture arena. The overlaid sequence shows the measured flight.}
  \label{fig:hardware-validation}
\end{figure*}

All three logged runs entered the configured 3-D goal region. No planner update was rejected and no deadline was missed. For the representative run in Fig.~\ref{fig:hardware-validation}, QP solve time on the Dell Precision 3660 was 23.8~ms at the median, 26.5~ms at the 95th percentile, and 45.9~ms at the maximum; all values are below the 100~ms replanning period. RMS and maximum position tracking errors were 1.72 and 2.92~cm. Minimum planned obstacle clearance was 7.24~cm.

\section{Discussion}
\label{sec:discussion}

The controlled comparison reveals a consistent tradeoff. Across all 27 same passage starts, harmonic guidance increases minimum clearance and reduces RMS jerk, but lengthens the path. Both methods cross the same designated opening and use the same downstream planner, so coarse passage choice cannot by itself explain the ordering. 
The geometry of the guidance fields offers one possible explanation. Harmonic guidance comes from a smooth boundary value problem over the connected free space, while Dijkstra guidance follows accumulated graph cost. The harmonic field has no explicit shortest path objective. That difference is consistent with the observed combination of larger clearance and longer travel. Smooth spatial changes in the harmonic field may also produce more gradual changes in the queried directions, consistent with lower RMS jerk under the planner. The explanation is interpretive, not a theoretical guarantee. Harmonic smoothness alone does not imply larger clearance or lower jerk.

The maze and flight experiments test a separate question: whether short horizon planning can use guidance computed over the full environment. The 0.60~s horizon covers only a small fraction of each maze route, yet every query comes from a field solved over the global free space. Crazyflie trials execute the same pipeline on the physical vehicle: field query, constrained local planning, and reference tracking. The two experiments show that a short horizon planner can use globally constructed harmonic guidance over route scales far beyond one prediction horizon.

The present study assumes a known static environment, a fixed goal, and a third-order translational planning model. The planner uses only normalized harmonic directions along its nominal prediction; the scalar value and much of the surrounding field remain unused. Future work can combine $V$ and $\nabla V$, query the field at multiple spatial scales, and adapt the horizon or objective weights to local field geometry. Online field updates could extend the approach to changing goals or maps.

\section{Conclusion}
\label{sec:conclusion}

We demonstrated volumetric harmonic field navigation on a physical quadrotor using constrained predictive motion generation over a short horizon. Controlled tests showed a consistent tradeoff: harmonic guidance gave larger minimum clearance and lower RMS jerk, but longer paths; the same pattern remained after coarse passage choice was fixed. Long maze tests and Crazyflie flights showed that globally constructed harmonic guidance can support local predictive motion over route scales far beyond one horizon and on a physical platform. The results establish volumetric harmonic guidance as a viable navigation approach for quadrotors in known cluttered environments.

\balance
\bibliographystyle{IEEEtran}
\bibliography{references}
\end{document}